\documentclass[journal]{IEEEtran}
\usepackage{graphicx}
\usepackage{amsmath,amssymb}
\usepackage{booktabs}
\usepackage[dvipsnames]{xcolor}
\usepackage{microtype}
\usepackage{xspace}
\usepackage{etoolbox}
\usepackage{capt-of}
\usepackage{enumitem}
\usepackage[numbers,sort&compress]{natbib}
\usepackage[hyphens]{url}
\setlist[itemize]{topsep=2pt,itemsep=1pt,parsep=0pt,leftmargin=*}
\definecolor{cvprblue}{rgb}{0.21,0.49,0.74}
\usepackage[breaklinks,colorlinks,allcolors=cvprblue]{hyperref}

\title{PhyVGGT: Feed-Forward Articulated Object Reconstruction from Monocular Videos\vspace{8pt}}
\author{Qi~Shao, Zhihao~Cao, Shuhao~Zhai, Anh~Nguyen, Hong~Liang, Liang~Li, and~Baoru~Huang%
\thanks{Corresponding authors: Liang Li and Baoru Huang.}%
\thanks{Q. Shao is with Harbin Engineering University, Harbin, China, and also with the University of Liverpool, Liverpool, U.K.}%
\thanks{Z. Cao is with ETH Zurich, Zurich, Switzerland, and also with the National University of Singapore, Singapore.}%
\thanks{S. Zhai is with the University of Macau, Macau, China.}%
\thanks{A. Nguyen and B. Huang are with the University of Liverpool, Liverpool, U.K., and also with Imperial College London, London, U.K.}%
\thanks{H. Liang and L. Li are with Harbin Engineering University, Harbin, China.}}
\makeatletter
\let\phy@maketitle\@maketitle
\renewcommand{\@maketitle}{\phy@maketitle
\vspace{12pt}\setcounter{figure}{0}%
\begin{center}
\centering
\includegraphics[width=\textwidth]{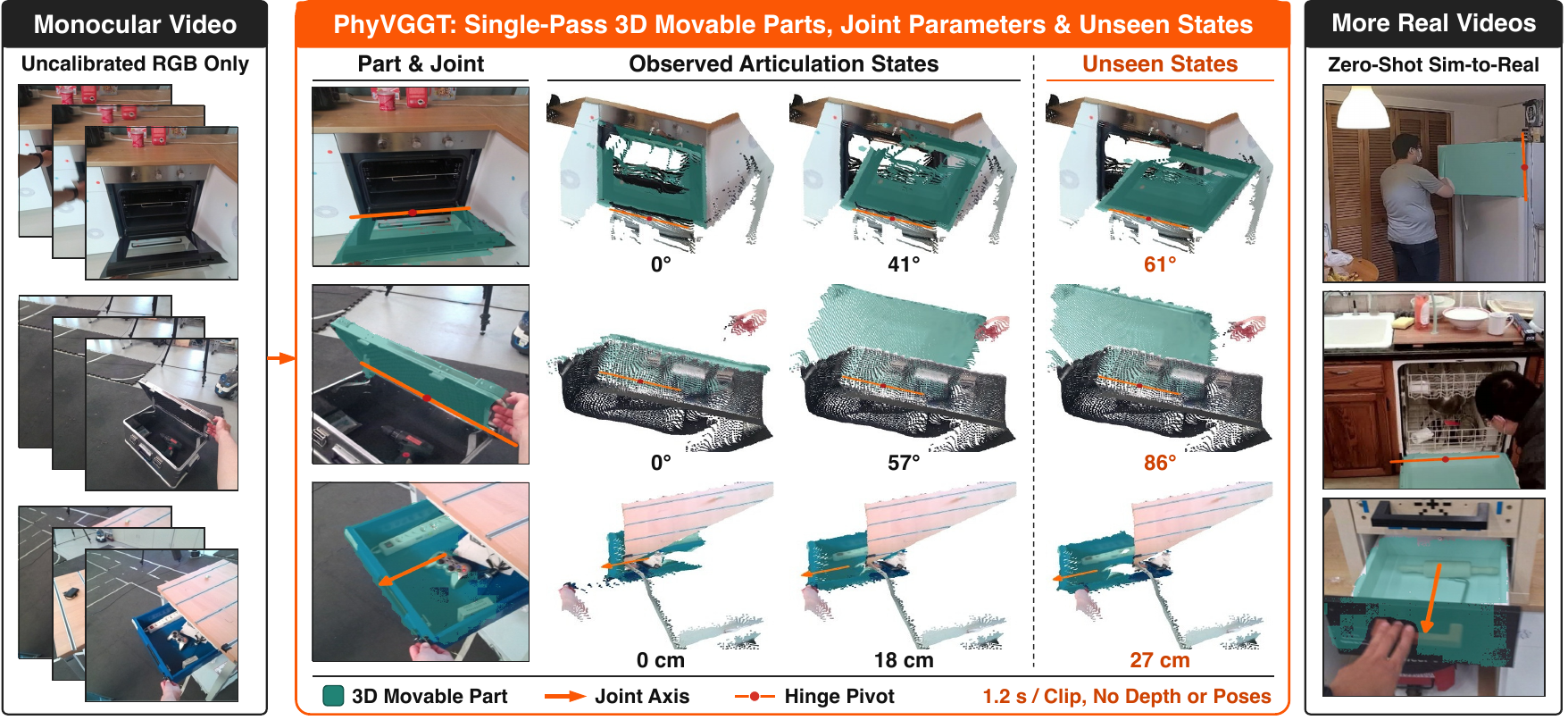}
\vspace{-0.3em}
\captionof{figure}{\textbf{Overview of single-pass 3D articulated object reconstruction from uncalibrated monocular RGB videos.} Given casual RGB frames (\emph{left}), PhyVGGT segments 3D movable parts (teal), estimates joint axes (orange) and hinge pivots (red) across observed articulation states, and drives each part to unseen states in 1.2\,s per clip (\emph{centre}), generalising zero-shot from synthetic training to diverse real-world scenes (\emph{right}).}
\label{fig:teaser}
\end{center}%
\vspace{-0.2em}}
\makeatother
\begin{document}
\maketitle
\begin{abstract}
Reconstructing an articulated object with its 3D movable parts, joint parameters and unseen articulation states from an uncalibrated monocular video is essential for robot manipulation and interactive digital twins.
Existing per-scene optimisation methods require minutes to hours and external depth or camera poses, whereas feed-forward 3D foundation models reconstruct dense point maps in one pass without segmenting movable parts or inferring joints.
Even on these monocular point maps, regressing or fitting continuous joints directly fails on casual videos, since cross-frame attention blurs moving part boundaries in deep layers, underestimates motion along the viewing ray and leaves continuous axes ill-conditioned under small opening angles.
To address these challenges, we propose PhyVGGT, a single-pass framework on a frozen 3D foundation model that anchors moving-part matching and closed-form joint estimation to the undistorted geometry of the stationary base.
First, our Part and Motion Decoder fuses intermediate per-frame tokens with global tokens and references part queries to a static-base query to segment 3D parts, align point maps on the static base and predict learned Bingham priors.
Second, our Kinematic Inference Layer fits an orthogonal Manhattan frame to static surface normals, selects the joint axis in closed form under a Bingham posterior combining pairwise motion scatter with the learned priors, and solves a linear system for the hinge pivot to drive unseen states.
Trained solely on synthetic data, PhyVGGT achieves state-of-the-art monocular performance across five benchmarks, outperforming FAMOS and monocular ArtiPoint, surpassing posed RGB-D ArtiPoint in median error, and running in 6.8\,s per 48-frame video against hours of per-scene optimisation in Articulat3D.
Project page: \url{https://phyvggt.github.io}.
\end{abstract}
\begin{IEEEkeywords}
Articulated object reconstruction, joint estimation, 3D foundation models, monocular video, feed-forward reconstruction.
\end{IEEEkeywords}

\section{Introduction}
\label{sec:intro}

Robotic manipulation and interactive simulation require reconstructing an articulated object in 3D, including its geometry, movable parts, joint types, axes, pivots and joint states. Although feed-forward 3D foundation models~\cite{dust3r,vggt,pi3} recover dense point maps and cameras from uncalibrated images in one pass and token query heads segment static spatial relations~\cite{relationvggt}, neither models how object parts move. Conversely, per-scene optimisation methods~\cite{paris,artgs,videoartgs,articulat3d} fit explicit motion models to multi-state scans or videos, yet require minutes to hours per object alongside sensor depth, camera poses or known kinematic trees. While recent feed-forward articulation predictors~\cite{artsplat,famos,sim2art} avoid per-scene fitting, they regress unconstrained joint vectors or fit continuous displacements directly in 3D, where monocular depth errors corrupt the recovered axis and pivot.

Attaching a prediction head or motion fitter directly to a frozen 3D foundation model fails on casual videos because alternating intra- and cross-frame attention~\cite{vggt} treats the scene as static. In deep global-attention layers, moving part edges blur into the stationary body and camera drift mixes with part motion, while cross-frame context biases moving-part depth towards other frames and underestimates displacement along the viewing ray. When the canonical rest pose is also unknown, Fisher information for continuous axis and pivot estimation vanishes quadratically under small opening angles~\cite{woltring1985}, leaving unconstrained regression and motion fitting ill-conditioned.

Unlike moving parts, the stationary base is unaffected by cross-frame depth bias and remains geometrically accurate across frames. In manufactured cabinets, appliances and furniture, flat faces meet at right angles, so revolute and prismatic axes align with the three orthogonal directions of the static base's Manhattan frame~\cite{straub2014mixture} while hinge pivots lie along the boundary between the part and the base. Benefiting from this undistorted anchor, we redesign both feature decoding and joint inference around the stationary base. Referencing part queries to the static base cancels camera drift during dense matching, and restricting candidate axes to the static Manhattan frame turns ill-conditioned continuous regression into closed-form three-way selection that tolerates up to $45^{\circ}$ of tilt towards the viewing ray.

Based on this design, we propose PhyVGGT, a single-pass framework on a frozen visual geometry model~\cite{vggt} that reconstructs 3D movable parts, joint parameters and unseen articulation states from an uncalibrated monocular video, as shown in Figure~\ref{fig:teaser}. First, our Part and Motion Decoder introduces multi-layer per-frame and global token fusion together with an explicit static-base query, recovering sharp 3D part masks, correspondences aligned to the static base and learned Bingham priors. Second, our Kinematic Inference Layer fits an orthogonal Manhattan frame to static surface normals via fourth-order alignment, selects the joint axis in closed form under a Bingham posterior~\cite{bingham1974} combining pairwise motion scatter with the learned priors, and solves a positive-definite linear system for the hinge pivot to drive unseen states.

Our contributions are as follows:
\begin{itemize}
\item We introduce PhyVGGT, a single-pass framework on a frozen 3D model that anchors part matching and closed-form joint estimation to the stationary base to reconstruct 3D parts, joints and unseen states from a monocular video.
\item We propose a Part and Motion Decoder that fuses multi-layer per-frame and global tokens and references part queries to a static-base query to recover sharp 3D part masks and correspondences aligned on the static base.
\item We design a Kinematic Inference Layer that restricts joint estimation to the Manhattan frame of the static base, removing depth bias along the viewing ray through closed-form Bingham axis selection and linear hinge pivot estimation.
\end{itemize}

\section{Related Work}
\label{sec:related}

\subsection{Articulated Object Reconstruction}
Early learning methods predict part coordinates and joint parameters from depth scans~\cite{ancsh,screwnet}, whereas per-scene optimisation fits neural radiance fields or 3D Gaussians to discrete articulation states~\cite{paris,artgs} or continuous videos~\cite{videoartgs,articulat3d}. Articulation in Prime~\cite{aip} shortens per-object fitting to roughly ten minutes, yet still requires external depth maps, camera poses and part masks. Casual and egocentric pipelines likewise depend on RGB-D input~\cite{itaco}: ArtiPoint~\cite{artipoint} fits joints to tracked 3D interaction points under known odometry, MoMa-SG~\cite{momasg} constructs articulated scene graphs from RGB-D sequences, and PAWS~\cite{paws} lifts 2D hand trajectories with sensor depth and poses before querying a vision-language model over surface candidates. Avoiding per-scene optimisation and depth sensors, single-image predictors infer openable parts from appearance priors~\cite{opdmulti,monoart,urdfanything,articulateanything}, ArtSplat~\cite{artsplat} and FAMOS~\cite{famos} predict joints from multi-state point clouds, and sim2art~\cite{sim2art} estimates part masks and continuous joint vectors from monocular videos. Yet predicting unconstrained vectors in $\mathbb{R}^3$ or fitting continuous axes directly to monocular point maps propagates depth errors along the viewing ray into the joint estimate, whereas PhyVGGT anchors part matching and closed-form Bingham inference to the Manhattan frame of the static base in one RGB pass.

\subsection{Feed-Forward 3D Foundation Models}
DUSt3R~\cite{dust3r} and MASt3R~\cite{mast3r} predict dense 3D point maps from uncalibrated image pairs in a shared coordinate frame, while VGGT~\cite{vggt} and $\pi^3$~\cite{pi3} recover point maps, camera poses and point tracks across an entire video in one forward pass. Recent extensions adapt visual geometry transformers to causal streaming and dynamic 4D scenes~\cite{monst3r,st4rtrack,streamvggt,stream3r}, and such reconstruction priors also support monocular dense mapping across multiple agents~\cite{como3r,mags2025}. RelationVGGT~\cite{relationvggt} attaches a lightweight query decoder to a frozen VGGT backbone for static 3D spatial relation segmentation. Although these foundation models reconstruct accurate stationary 3D surfaces in a single pass, their alternating intra- and cross-frame attention blurs moving part boundaries in deep layers and biases moving-part depth towards context frames, underestimating motion along the viewing ray. PhyVGGT overcomes both issues by fusing intermediate per-frame and global tokens relative to a static-base query and anchoring joint inference to undistorted static surface normals.

\section{Method}
\label{sec:method}

\begin{figure*}[t]
\centering
\includegraphics[width=\textwidth]{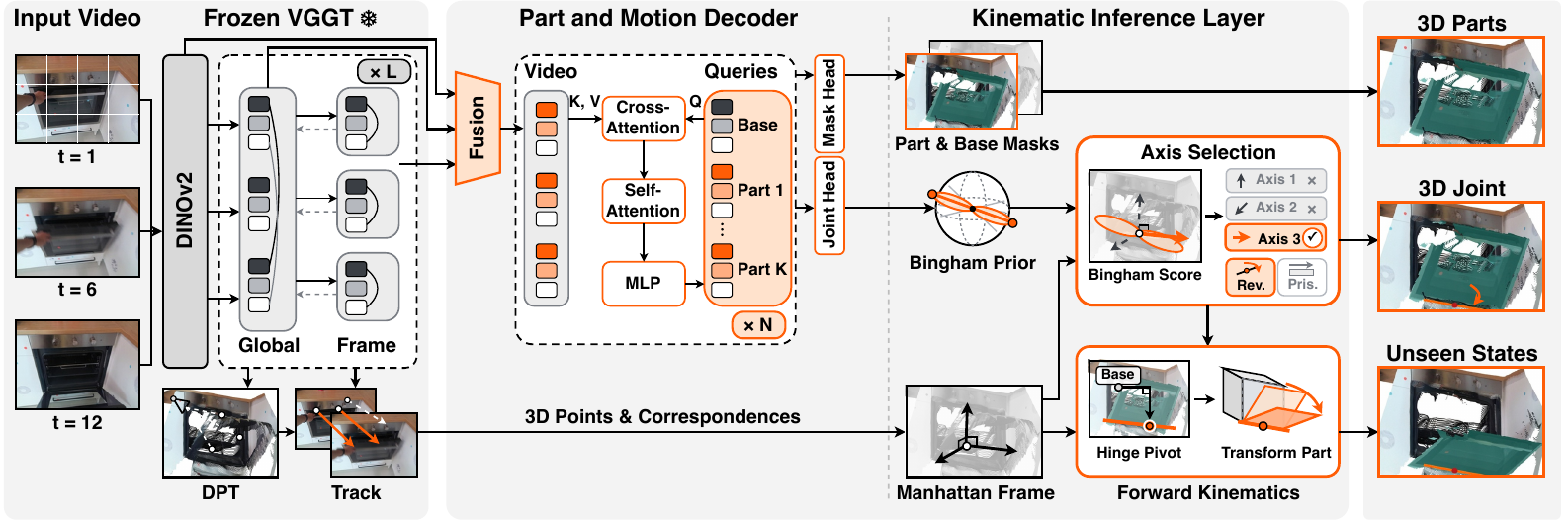}
\caption{\textbf{Architecture overview of PhyVGGT.} Given an uncalibrated monocular video, a frozen VGGT backbone extracts patch, per-frame and global tokens alongside 3D point maps (\emph{DPT}) and tracking features (\emph{Track}). First, our trainable Part and Motion Decoder fuses these streams and updates learnable base and part queries across $N$ attention blocks to output part masks (\emph{Mask Head}), correspondences and Bingham priors (\emph{Joint Head}). Second, our Kinematic Inference Layer fits a \emph{Manhattan Frame} to static normals, scores its three axes via a three-way softmax (\emph{Axis Selection}) and solves for joint type, \emph{Hinge Pivot} and \emph{Unseen States} via \emph{Forward Kinematics}.}
\label{fig:pipeline}
\end{figure*}

Given an uncalibrated monocular RGB video $\{\mathbf{I}_t\}_{t=1}^{T}$, we aim to reconstruct an articulated object consisting of a stationary base $\mathrm{b}$ and up to $K$ movable rigid parts in a single forward pass. Each movable part $k$ connects to the base via a one-degree-of-freedom joint defined by its kinematic type $y_k \in \{\text{rev}, \text{pris}\}$, unsigned unit axis $\mathbf{a}_k \in \mathbb{R}\mathbb{P}^2$, hinge pivot $\mathbf{o}_k \in \mathbb{R}^3$ and per-frame joint states $\{q_{kt}\}_{t=1}^{T}$. Together, these parameters specify a unit twist $\boldsymbol{\xi}_k$, equal to $(\mathbf{a}_k, \mathbf{o}_k \times \mathbf{a}_k)$ for a revolute joint and $(\mathbf{0}, \mathbf{a}_k)$ for a prismatic joint, which transforms any homogeneous 3D point $\mathbf{x}$ on part $k$ from frame $1$ to frame $t$:
\begin{equation}
\mathbf{x}_t = \exp\!\big(q_{kt}\, \hat{\boldsymbol{\xi}}_k\big)\, \mathbf{x}, \qquad q_{k1} = 0,
\label{eq:screw}
\end{equation}
where $\exp(q_{kt}\hat{\boldsymbol{\xi}}_k) \in SE(3)$ is generated by $\hat{\boldsymbol{\xi}}_k \in \mathfrak{se}(3)$. From RGB frames alone, a frozen VGGT~\cite{vggt} backbone predicts world-frame point maps $\{\mathbf{X}_t\}_{t=1}^{T}$, tracking features and multi-layer tokens. Benefiting from the undistorted stationary surfaces in $\{\mathbf{X}_t\}_{t=1}^{T}$, PhyVGGT anchors two sequential modules to the static base as shown in Figure~\ref{fig:pipeline}: our Part and Motion Decoder in Sec.~\ref{sec:decoder} segments 3D parts and recovers rigid part motion relative to the static base, and our Kinematic Inference Layer in Sec.~\ref{sec:bingham} selects joint axes on the Manhattan frame of the static base and solves for pivots and states.

\subsection{Part and Motion Decoder}
\label{sec:decoder}

In VGGT's alternating aggregator, per-frame self-attention preserves sharp intra-frame part edges, whereas deep cross-frame global attention captures inter-frame motion while smoothing moving boundaries into the static body. To retain both cues, our Part and Motion Decoder extracts per-frame and global tokens from multiple VGGT layers and fuses them with DINOv2~\cite{dinov2} patch tokens $\mathbf{D}_t$ and VGGT tracking features to form per-frame features $\mathbf{U}^{\mathrm{f}}_t$ and global features $\mathbf{U}^{\mathrm{g}}_t$. To decode parts relative to the static body, we maintain $K + 2$ learnable queries $\mathbf{Q} = [\mathbf{q}_{\mathrm{bg}}, \mathbf{q}_{\mathrm{b}}, \mathbf{q}_1, \dots, \mathbf{q}_K]$ for background, the static base and $K$ movable parts, replicated per frame as $\mathbf{Q}_t$. In each of the $N$ residual decoder blocks shown in the middle of Figure~\ref{fig:pipeline}, $\mathbf{Q}_t + \mathbf{Q}$ first cross-attends to keys and values from $\mathbf{U}^{\mathrm{f}}_t$ within frame $t$ to localise sharp part boundaries, followed by a feed-forward layer. Clip queries $\mathbf{Q}$ then add the temporal mean of $\{\mathbf{Q}_t\}$, cross-attend to keys and values from $\{\mathbf{U}^{\mathrm{g}}_t\}_{t=1}^{T}$ across frames, separate part instances via self-attention, and update each part query $\mathbf{q}_k$ through an MLP on $[\mathbf{q}_k, \mathbf{q}_{\mathrm{b}}, \mathbf{q}_k - \mathbf{q}_{\mathrm{b}}]$ before adding $\mathbf{Q}$ back to $\mathbf{Q}_t$.

From the decoded features, upsampled patch tokens are combined with VGGT point-map, tracking and image features into pixel embeddings $\mathbf{U}_t(u) \in \mathbb{R}^{d}$ and coarse mask logits $\mathbf{c}_t(u)$. Projecting $\mathbf{Q}_t$ to $\tilde{\mathbf{Q}}_t \in \mathbb{R}^{(K+2)\times d}$ yields pixel mask probabilities:
\begin{equation}
\mathbf{P}_t(u) = \mathrm{softmax}\!\big(\mathbf{c}_t(u) + d^{-1/2}\tilde{\mathbf{Q}}_t\, \mathbf{U}_t(u)\big).
\label{eq:mask}
\end{equation}
In parallel, a matching head maps tracking features, query-conditioned tokens and patch tokens $\mathbf{D}_t$ to a unit pixel descriptor $\mathbf{d}_t(u)$. With temperature $\tau > 0$, weighting descriptor similarity by $\mathbf{P}_t(u)\cdot\mathbf{P}_r(v)$, the probability that pixel $u$ in frame $t$ and pixel $v$ in reference frame $r$ belong to the same slot, restricts correspondences within each rigid part:
\begin{equation}
a^r_t(u, v) = \frac{\exp(\tau^{-1}\mathbf{d}_t(u)^{\top}\mathbf{d}_r(v))\,\mathbf{P}_t(u)\!\cdot\!\mathbf{P}_r(v)}{\textstyle\sum_m \exp(\tau^{-1}\mathbf{d}_t(u)^{\top}\mathbf{d}_r(m))\,\mathbf{P}_t(u)\!\cdot\!\mathbf{P}_r(m)},
\label{eq:match}
\end{equation}
which maps pixel $u$ to its matched 3D point $\mathbf{C}^r_t(u) = \sum_v a^r_t(u, v)\mathbf{X}_r(v)$ in frame $r$. Using static-base correspondences $\mathbf{C}^1_t(u) \leftrightarrow \mathbf{X}_t(u)$, iteratively reweighted orthogonal Procrustes alignment first estimates a rigid transformation per frame to align $\{\mathbf{X}_t\}_{t=1}^{T}$ on the static base and cancel camera drift. On these aligned point maps, applying the same Procrustes solver to each part's correspondences between frames $r$ and $t$ recovers its pairwise rotation $\mathbf{R}^r_{kt} \in SO(3)$, rotation vector $\mathbf{z}^{r,\mathrm{rev}}_{kt} = \frac{1}{2}(\mathbf{R}^r_{kt} - \mathbf{R}^{r\top}_{kt})^{\vee}$ and centroid translation $\mathbf{z}^{r,\mathrm{pri}}_{kt} = \bar{\mathbf{x}}_{kt} - \bar{\mathbf{c}}^r_{kt}$ between matched centroids $\bar{\mathbf{c}}^r_{kt}$ and $\bar{\mathbf{x}}_{kt}$, along with total inlier weight $\omega^r_{kt}$. Finally, the joint head attends across features along the boundary between the part and the base alongside pairwise motion summaries to predict joint-type logits, a boundary pivot prior $\hat{\mathbf{o}}_k \in \mathbb{R}^3$, and Cholesky factors $\mathbf{L}_k$ forming positive-semidefinite Bingham prior matrices $\mathbf{M}_k = \mathbf{L}_k\mathbf{L}_k^{\top}$ for $\mathbf{M}_k \in \{\mathbf{M}^{\mathrm{rev}}_k, \mathbf{M}^{\mathrm{pri}}_k\}$.

\subsection{Kinematic Inference Layer}
\label{sec:bingham}

\begin{figure}[t]
\centering
\includegraphics[width=\columnwidth]{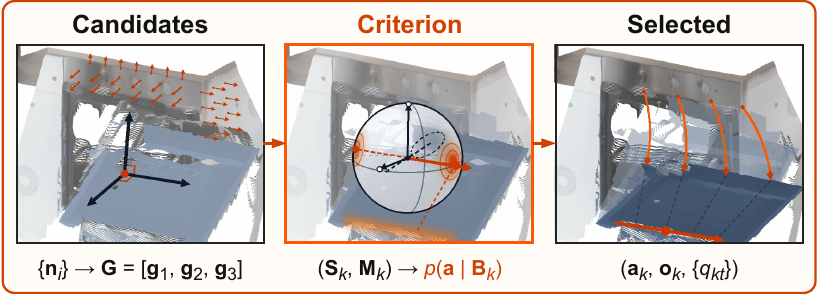}
\caption{\textbf{Mechanism of the Kinematic Inference Layer on real point maps.} Dense static-base normals $\{\mathbf{n}_i\}$ form an orthogonal Manhattan frame $\mathbf{G}=[\mathbf{g}_1,\mathbf{g}_2,\mathbf{g}_3]$ (\emph{Candidates}). Combining pairwise motion scatter $\mathbf{S}_k$ with the learned Bingham prior $\mathbf{M}_k$ defines an antipodal Bingham posterior $p(\mathbf{a}\mid\mathbf{B}_k)$ (\emph{Criterion}), selecting $\mathbf{a}_k$ to solve for the hinge pivot $\mathbf{o}_k$ and states $\{q_{kt}\}$ (\emph{Selected}).}
\label{fig:design_bingham}
\end{figure}

Given the pairwise motion vectors $\mathbf{z}^r_{kt} \in \{\mathbf{z}^{r,\mathrm{rev}}_{kt}, \mathbf{z}^{r,\mathrm{pri}}_{kt}\}$ and inlier weights $\omega^r_{kt}$ from the decoder, summing outer products over all frame pairs cancels single-frame reference noise and forms the rotational and translational motion scatter matrices:
\begin{equation}
\mathbf{S}_k = \sum_{r,t} \omega^r_{kt}\, \mathbf{z}^r_{kt} \mathbf{z}^{r\top}_{kt}.
\label{eq:scatter}
\end{equation}
Under noise-free 3D geometry, $\mathbf{S}_k$ has rank one along $\mathbf{a}_k$. On monocular point maps, cross-frame attention scales motion along the viewing ray by $\gamma \in (0, 1)$ and tilts the leading eigenvector of $\mathbf{S}_k$ by up to $\tan\theta \le (1-\gamma)/(2\sqrt{\gamma})$ regardless of frame count, while small opening angles under an unknown rest pose leave continuous axis estimation ill-conditioned, as derived in the supplementary material. Instead of estimating an unconstrained direction on $\mathbb{R}\mathbb{P}^2$, selecting $\mathbf{a}_k$ from three orthogonal candidate axes in Figure~\ref{fig:design_bingham} recovers the exact direction whenever this tilt stays below $45^{\circ}$, which holds for any $\gamma > 3 - 2\sqrt{2} \approx 0.172$.

To obtain these orthogonal candidates without depth bias from moving parts, we compute finite-difference unit normals $\{\mathbf{n}_i\}_{i \in \mathcal{O}}$ on the point-map grid of the stationary object regions. In manufactured cabinets, appliances and furniture, planar faces meet at right angles~\cite{straub2014mixture,real2code}, so $\{\mathbf{n}_i\}$ cluster around an orthogonal Manhattan frame $\mathbf{G} = [\mathbf{g}_1, \mathbf{g}_2, \mathbf{g}_3] \in SO(3)$. Whereas point-cloud PCA rotates arbitrarily when two object dimensions are similar, maximising fourth-order normal alignment~\cite{zhai2020complete} locks directly onto the perpendicular face normals on the left of Figure~\ref{fig:design_bingham}:
\begin{equation}
\mathbf{G} = \arg\max_{\mathbf{G} \in SO(3)} \sum_{i \in \mathcal{O}} \sum_{j=1}^{3} \big(\mathbf{n}_i^{\top} \mathbf{g}_j\big)^4,
\label{eq:gauge}
\end{equation}
which we solve from second-moment eigenvectors by projected gradient ascent on $SO(3)$ via polar decomposition.

Since $\pm\mathbf{a}_k$ define the same physical joint line, we combine the motion scatter $\mathbf{S}_k$ and the learned prior $\mathbf{M}_k \in \{\mathbf{M}^{\mathrm{rev}}_k, \mathbf{M}^{\mathrm{pri}}_k\}$ under an antipodally symmetric Bingham distribution~\cite{bingham1974} $p(\mathbf{a}\mid\mathbf{B}_k) \propto \exp(\mathbf{a}^{\top} \mathbf{B}_k \mathbf{a})$ on $\mathbb{R}\mathbb{P}^2$. Here $\alpha_k > 0$ normalises $\mathbf{S}_k$ by total inlier weight and point variance, multiplied by the part's squared radius of gyration for revolute scatter so both joint types are dimensionless, while $\kappa_k > 0$ weights $\mathbf{M}_k$ and increases when translational motion aligns with the viewing ray:
\begin{equation}
\mathbf{B}_k = \alpha_k\, \mathbf{S}_k + \kappa_k\, \mathbf{M}_k.
\label{eq:bmat}
\end{equation}
Although normalising $p(\mathbf{a}\mid\mathbf{B}_k)$ over $\mathbb{S}^2$ requires confluent hypergeometric functions, restricting $\mathbf{a}_k$ to the three columns of $\mathbf{G}$ with inverse temperature $\beta > 0$ reduces exact posterior inference to the closed-form three-way softmax in the middle of Figure~\ref{fig:design_bingham}:
\begin{equation}
w_{kj} = \frac{\exp\!\big(\beta\, \mathbf{g}_j^{\top} \mathbf{B}_k\, \mathbf{g}_j\big)}{\sum_{m=1}^{3} \exp\!\big(\beta\, \mathbf{g}_m^{\top} \mathbf{B}_k\, \mathbf{g}_m\big)}, \qquad j \in \{1,2,3\}.
\label{eq:bingham}
\end{equation}
Orienting each candidate by $s_{kj} = \mathrm{sign}(\mathbf{g}_j^{\top}\hat{\mathbf{a}}_k)$ using the leading eigenvector $\hat{\mathbf{a}}_k$ of $\mathbf{M}_k$, we select $j^* = \arg\max_j w_{kj}$ with straight-through gradients during training:
\begin{equation}
\mathbf{a}_k = s_{k j^*}\, \mathbf{g}_{j^*}.
\label{eq:axis}
\end{equation}

Given $\mathbf{a}_k$, we infer the joint type $y_k$ by combining the joint head's type logits with the reduction in Procrustes residual from pure translation to rigid rotation, penalising lever arms larger than the object radius. For a revolute joint with pairwise translation residual $\mathbf{t}^r_{kt} = \bar{\mathbf{x}}_{kt} - \mathbf{R}^r_{kt}\bar{\mathbf{c}}^r_{kt}$, normalised pair weight $\bar{\omega}^r_{kt}$ and projected rotation operator $\mathbf{A}^r_{kt} = (\mathbf{I} - \mathbf{R}^r_{kt})(\mathbf{I} - \mathbf{a}_k\mathbf{a}_k^{\top})$, we estimate the hinge pivot $\mathbf{o}_k = \arg\min_{\mathbf{o}} E_k(\mathbf{o})$ by minimising:
\begin{equation}
E_k(\mathbf{o}) = \sum_{r,t} \bar{\omega}^r_{kt}\big\|\mathbf{A}^r_{kt}\mathbf{o} - \mathbf{t}^r_{kt}\big\|^2 + \lambda_{\mathrm{o}}\|\mathbf{o} - \hat{\mathbf{o}}_k\|^2.
\label{eq:pivot}
\end{equation}
Setting $\nabla_{\mathbf{o}}E_k(\mathbf{o}) = \mathbf{0}$ yields a $3\times 3$ positive-definite linear system that solves for the rotation centre in the plane orthogonal to $\mathbf{a}_k$ from rigid motion while fixing $\mathbf{a}_k^{\top}\mathbf{o}_k = \mathbf{a}_k^{\top}\hat{\mathbf{o}}_k$ along the axis to the boundary prior $\hat{\mathbf{o}}_k$ on the right of Figure~\ref{fig:design_bingham}. Projecting frame-$1$-to-$t$ rotation angles or centroid translations onto $\mathbf{a}_k$ recovers $\{q_{kt}\}_{t=1}^{T}$, and substituting $(\mathbf{a}_k, \mathbf{o}_k, q_{kt})$ into Eq.~\eqref{eq:screw} transforms the 3D part points to unseen states in one pass.

\subsection{Training Objective}
\label{sec:training}

Keeping the VGGT backbone frozen, we train PhyVGGT in two stages on synthetic clips, first optimising the segmentation and correspondence trunk and then training the joint priors and calibration parameters of the Kinematic Inference Layer. Predicted part slots are matched to ground-truth parts via Hungarian assignment on pixel mask overlap. After $\mathrm{Sim}(3)$ scene alignment, the first stage sums cross-entropy and Dice losses on $\mathbf{P}_t$, first-frame InfoNCE on $\mathbf{d}_t(u)$ and 3D canonical correspondence losses on $\mathbf{C}^1_t(u)$, while the second stage sums binary cross-entropy on joint-type logits, Bingham quadratic and antipodal cosine losses on joint axes, smooth-$L_1$ losses on per-frame screw motion vectors and motion span, and a point-to-line distance loss on $\mathbf{o}_k$. All training clips are rendered from PartNet-Mobility~\cite{sapien} with moving cameras, textured indoor rooms, articulated hands grasping handles and hinged doors converted into sliding doors, supervising 3D motion-based joint estimation without real-world videos.

\begin{figure*}[t]
\centering
\includegraphics[width=\textwidth]{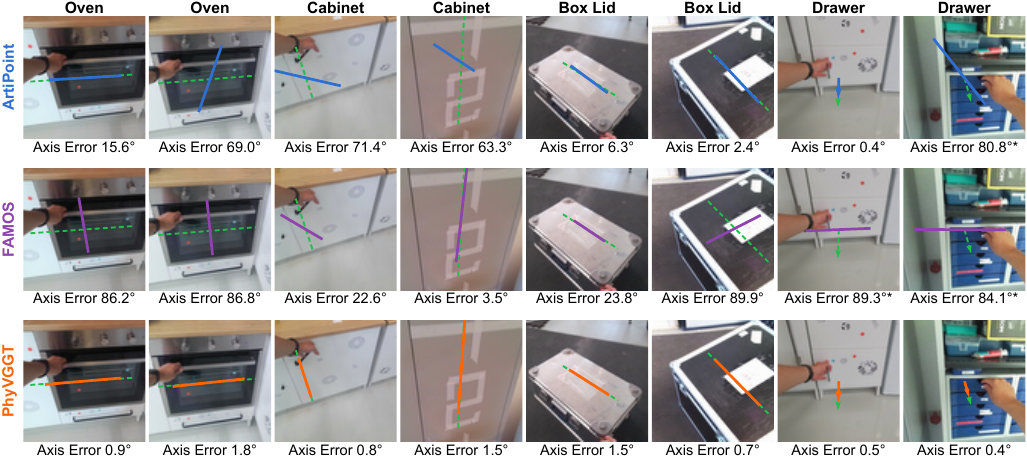}
\caption{\textbf{Qualitative comparison of 3D joint axis estimation on Arti4D test videos.} Blue shows ArtiPoint with RGB-D input and ground-truth poses, purple shows FAMOS on PhyVGGT point maps, orange shows RGB-only PhyVGGT, and green dashed lines denote ground truth. Arrows indicate prismatic joints while lines indicate revolute joints, with $^{*}$ marking wrong joint types.}
\label{fig:cmp_real}
\end{figure*}

\section{Experiments}
\label{sec:experiments}

\subsection{Experimental Setup}
\label{sec:setup}

\noindent\textbf{Datasets.} We evaluate zero-shot synthetic-to-real transfer across three real-world video benchmarks alongside in-domain and cross-category generalisation on two synthetic video benchmarks. On real-world interactions, Arti4D~\cite{artipoint} captures egocentric handheld RGB-D videos of drawers, doors and appliances across 329 held-out test clips in three unseen scenes, D3D-HOI~\cite{d3dhoi} provides 170 third-person static-camera test clips annotated with 3D joint axes, and RBO~\cite{rbo} contributes 357 third-person RGB-D videos across 14 objects with motion-capture ground truth. On synthetic PartNet-Mobility~\cite{sapien} videos, we train on 6,264 clips from ten categories and evaluate on held-out test splits of 290 joints from the ten trained categories (Syn-10) and 510 joints across 38 categories including 28 unseen categories (Syn-38), as well as 48-frame reconstruction across all 20 scenes and 48 joints of Articulat3D-Sim~\cite{articulat3d}. Component ablations and backbone transfer are evaluated on four disjoint development splits: Arti4D-dev on scene \emph{din080} with 85 clips, D3D-HOI-dev with 86 clips, Syn-10-val with 313 joints and Syn-38-val with 534 joints.

\noindent\textbf{Metrics.} Following~\cite{artipoint,videoartgs}, we measure axis error as the acute angle on $\mathbb{R}\mathbb{P}^2$ and revolute pivot error as 3D line-to-line distance, reporting median errors and joint accuracy within $5^{\circ}$ and $10^{\circ}$ requiring both correct joint type and axis.

\noindent\textbf{Baselines.} Among uncalibrated monocular RGB methods, we evaluate three complementary baseline groups: same-backbone ablations on frozen VGGT via \emph{Direct regression} with a continuous MLP joint head trained on the same features, masks and synthetic clips as PhyVGGT and \emph{Motion estimate} taking the leading eigenvector of $\mathbf{S}_k$ without Bingham selection on the Manhattan frame, appearance and point-cloud regressors via single-image OPDMulti~\cite{opdmulti} testing appearance priors without multi-frame 3D geometry and feed-forward FAMOS~\cite{famos} testing multi-state continuous point-cloud regression, and continuous trajectory fitting via monocular ArtiPoint~\cite{artipoint} on $\pi^3$~\cite{pi3} and VGGT~\cite{vggt}. Privileged baselines marked with $^{\ast}$ include posed RGB-D ArtiPoint~\cite{artipoint} with sensor depth and ground-truth poses, and per-scene 3D Gaussian optimisation Articulat3D~\cite{articulat3d} with ground-truth depth, poses, masks and kinematic trees.

\noindent\textbf{Implementation.} At test time, PhyVGGT processes $T = 12$ uniform $504 \times 504$ RGB frames per clip, and $T = 48$ on Articulat3D-Sim, using $K = 8$ part queries, $N = 3$ decoder blocks, 256-D queries, 64-D pixel and correspondence embeddings and intermediate VGGT layers $\{4, 11, 17, 23\}$. Reweighted Procrustes with a Cauchy loss runs three iterations for static-base alignment and six for part motion, while Eq.~\eqref{eq:pivot} solves a $3\times 3$ linear system with learned regulariser $\lambda_{\mathrm{o}} \ge 0.02$. Manhattan frame optimisation runs 48 projected gradient steps on a stride-8 normal grid, pooling background normals below 500 foreground normals or 50\% inliers and falling back to $\hat{\mathbf{a}}_k$ on untrusted frames deviating over $38^{\circ}$. We train on frozen VGGT-1B with AdamW using 8-frame synthetic clips of batch size 4 for 4,500 first-stage steps at learning rate $2\times 10^{-4}$ and 2,000 second-stage steps at $3\times 10^{-4}$. On real videos, person pixels are masked~\cite{deeplabv3} prior to matching, and on clips where a single part is operated the active slot is chosen automatically by motion amplitude weighted by root mask area. All runtimes are measured on a dedicated NVIDIA A100 GPU.

\subsection{Benchmark Comparisons}
\label{sec:exp_real}

\begin{table}[t]
\centering
\small
\caption{\textbf{Comparison on real-world Arti4D and D3D-HOI.} Joint accuracy and median axis error in degrees. $^{\ast}$Sensor depth and poses.}
\label{tab:real}
\setlength{\tabcolsep}{2.2pt}
\begin{tabular}{l ccc ccc}
\toprule
& \multicolumn{3}{c}{Arti4D (329)} & \multicolumn{3}{c}{D3D-HOI (170)} \\
\cmidrule(lr){2-4}\cmidrule(lr){5-7}
Method & $\le5^{\circ}$ & $\le10^{\circ}$ & Med. & $\le5^{\circ}$ & $\le10^{\circ}$ & Med. \\
\midrule
\textcolor{gray}{ArtiPoint, RGB-D$^{\ast}$} & \textcolor{gray}{0.422} & \textcolor{gray}{0.614} & \textcolor{gray}{6.30} & \textcolor{gray}{--} & \textcolor{gray}{--} & \textcolor{gray}{--} \\
\midrule
Direct regression & 0.006 & 0.020 & 57.07 & 0.022 & 0.063 & 37.15 \\
OPDMulti & 0.146 & 0.292 & 64.15 & 0.147 & 0.418 & 11.55 \\
ArtiPoint, $\pi^3$ & 0.198 & 0.389 & 15.90 & -- & -- & -- \\
FAMOS & 0.241 & 0.369 & 23.24 & 0.108 & 0.145 & 68.93 \\
Motion estimate & 0.090 & 0.280 & 21.92 & 0.186 & 0.369 & 15.93 \\
\textbf{PhyVGGT (Ours)} & \textbf{0.493} & \textbf{0.604} & \textbf{4.79} & \textbf{0.629} & \textbf{0.724} & \textbf{2.77} \\
\bottomrule
\end{tabular}
\end{table}

\begin{figure*}[t]
\centering
\includegraphics[width=\textwidth]{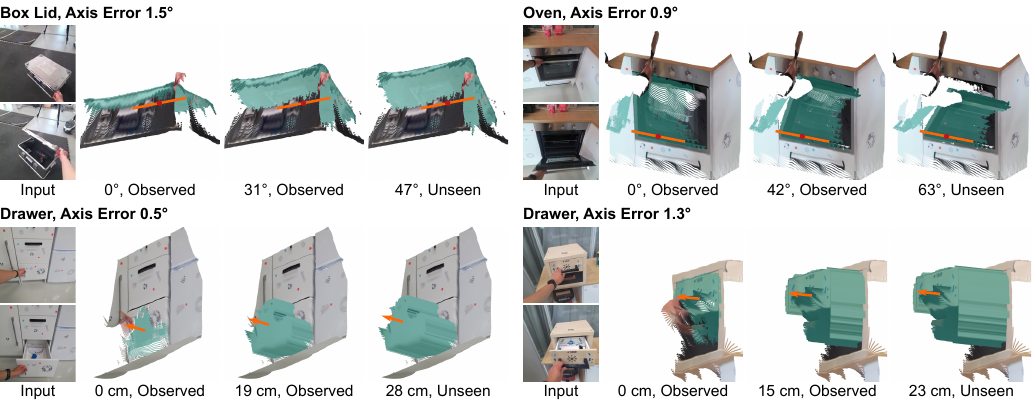}
\caption{\textbf{Qualitative 3D part reconstructions and unseen articulation states on Arti4D test videos.} For each scene, we show two input RGB frames alongside PhyVGGT's 3D point map near the rest state, in which the segmented \textbf{movable part (teal)} is driven by the predicted \textbf{joint axis (orange)} and \textbf{hinge pivot (red)} to the start and end of the observed motion and then to an unseen articulation state via forward kinematics. Drawers show the volume swept by the part along the predicted axis, with axis angular errors listed above.}
\label{fig:gallery}
\end{figure*}

Table~\ref{tab:real} evaluates zero-shot transfer from synthetic training to handheld egocentric videos on Arti4D and third-person videos on D3D-HOI. Benefiting from the undistorted geometry of stationary object faces, we redesigned axis estimation in our Kinematic Inference Layer around the object's Manhattan frame instead of camera coordinates, avoiding overfitting to synthetic camera viewpoints under steep handheld pitches and lifting $10^{\circ}$ accuracy to 60.4\% and 72.4\% against 2.0\% and 6.3\% for Direct regression and 29.2\% and 41.8\% for single-image OPDMulti. By introducing closed-form Bingham selection that fuses pairwise motion scatter with learned Bingham priors, the Kinematic Inference Layer also removes depth underestimation along the viewing ray and ambiguity under small opening angles, outperforming multi-state FAMOS at 60.4\% and 72.4\% against 36.9\% and 14.5\% and raising prismatic drawer accuracy to 48.2\% and 58.3\% against 14.7\% and 11.1\% for unconstrained Motion estimate. On Arti4D, referencing part queries to the static-base query and aligning static-base correspondences in our Part and Motion Decoder cancels handheld camera drift, improving accuracy to 60.4\% against 38.9\% for monocular ArtiPoint on $\pi^3$. Even without the sensor depth and ground-truth camera poses used by RGB-D ArtiPoint, fitting the Manhattan frame to flat static cabinet normals avoids depth sensor noise on small handles, raising $5^{\circ}$ accuracy to 49.3\% against 42.2\% and reducing median error to $4.79^{\circ}$ against $6.30^{\circ}$ in Figures~\ref{fig:cmp_real} and~\ref{fig:gallery}.

\begin{table}[t]
\centering
\small
\caption{\textbf{Comparison on real-world RBO.} Joint accuracy, median axis error in degrees, and revolute/prismatic accuracy at $10^{\circ}$. $^{\ast}$Sensor depth and poses.}
\label{tab:rbo}
\setlength{\tabcolsep}{2.6pt}
\begin{tabular}{l c c c c c}
\toprule
Method & $\le5^{\circ}$ & $\le10^{\circ}$ & Med. & Rev. & Pris. \\
\midrule
\textcolor{gray}{ArtiPoint, RGB-D$^{\ast}$} & \textcolor{gray}{0.148} & \textcolor{gray}{0.246} & \textcolor{gray}{51.3} & \textcolor{gray}{0.253} & \textcolor{gray}{0.229} \\
\midrule
Direct regression & 0.015 & 0.046 & 48.7 & 0.041 & 0.059 \\
OPDMulti & 0.050 & 0.129 & 44.7 & 0.126 & 0.135 \\
ArtiPoint, $\pi^3$ & 0.034 & 0.098 & 64.2 & 0.119 & 0.042 \\
FAMOS & 0.045 & 0.119 & 68.9 & 0.153 & 0.024 \\
Motion estimate & 0.063 & 0.185 & 29.2 & 0.206 & 0.128 \\
\textbf{PhyVGGT (Ours)} & \textbf{0.134} & \textbf{0.241} & \textbf{23.3} & \textbf{0.275} & \textbf{0.149} \\
\bottomrule
\end{tabular}
\end{table}

On tabletop interactions in Table~\ref{tab:rbo}, hands frequently occlude moving parts while perturbing the object pose. By separating static-base masks from moving parts and gating correspondences within each slot, our Part and Motion Decoder isolates unoccluded base regions for reliable fourth-order normal alignment, enabling PhyVGGT to reach 24.1\% overall on RBO and 31.1\% on stationary-base clips against 18.5\% for Motion estimate, 11.9\% for FAMOS and 9.8\% for monocular ArtiPoint. Even compared with privileged RGB-D ArtiPoint using sensor depth and motion-capture poses, anchoring revolute inference to planar base faces and boundary priors in our Kinematic Inference Layer raises revolute accuracy to 27.5\% against 25.3\% and halves median axis error to $23.3^{\circ}$ against $51.3^{\circ}$.

\begin{table}[t]
\centering
\small
\caption{\textbf{Comparison on synthetic PartNet-Mobility.} Joint accuracy and median axis error on 10 trained categories in Syn-10 and 38 categories with 28 unseen in Syn-38. $^{\ddagger}$Pretrained on openable furniture.}
\label{tab:syn}
\setlength{\tabcolsep}{3.4pt}
\begin{tabular}{l l ccc}
\toprule
Benchmark & Method & $\le5^{\circ}$ & $\le10^{\circ}$ & Med. \\
\midrule
Syn-10 & Direct regression & 0.061 & 0.198 & 20.60 \\
(290) & OPDMulti$^{\ddagger}$ & 0.028 & 0.072 & 75.03 \\
 & FAMOS & 0.426 & 0.529 & 7.08 \\
 & Motion estimate & 0.366 & 0.543 & 7.96 \\
 & \textbf{PhyVGGT (Ours)} & \textbf{0.762} & \textbf{0.778} & \textbf{1.53} \\
\midrule
Syn-38 & Direct regression & 0.042 & 0.128 & 33.40 \\
(510) & OPDMulti$^{\ddagger}$ & -- & -- & -- \\
 & FAMOS & 0.314 & 0.398 & 32.82 \\
 & Motion estimate & 0.259 & 0.410 & 15.76 \\
 & \textbf{PhyVGGT (Ours)} & \textbf{0.610} & \textbf{0.631} & \textbf{2.27} \\
\bottomrule
\end{tabular}
\end{table}

\begin{table}[t]
\centering
\small
\caption{\textbf{Comparison on 48-frame Articulat3D-Sim.} $^{\ast}$Per-scene optimisation with GT depth, poses, masks and kinematic trees.}
\label{tab:a3d}
\resizebox{\columnwidth}{!}{%
\setlength{\tabcolsep}{2.2pt}
\begin{tabular}{l c cc c c}
\toprule
& Trained & \multicolumn{2}{c}{All joints (48)} & Unseen & \\
\cmidrule(lr){2-2}\cmidrule(lr){3-4}\cmidrule(lr){5-5}
Method & Med. & $\le5^{\circ}$ & $\le10^{\circ}$ & $\le10^{\circ}$ & Time \\
\midrule
\textcolor{gray}{Articulat3D$^{\ast}$} & \textcolor{gray}{2.63} & \textcolor{gray}{0.479} & \textcolor{gray}{0.521} & \textcolor{gray}{0.316} & \textcolor{gray}{2.6--11.9\,h} \\
\midrule
Direct regression & 21.67 & 0.035 & 0.139 & 0.000 & 6.8\,s \\
Motion estimate & 11.79 & 0.215 & 0.347 & 0.158 & 6.8\,s \\
\textbf{PhyVGGT (Ours)} & \textbf{2.20} & \textbf{0.431} & \textbf{0.486} & \textbf{0.246} & \textbf{6.8\,s} \\
\bottomrule
\end{tabular}%
}
\end{table}

\begin{table}[t]
\centering
\small
\caption{\textbf{3D part, surface and hinge pivot reconstruction.} Upper rows report 3D movable / part IoU and surface error, and lower rows report held-out hinge median error in cm and accuracy within 10\,cm.}
\label{tab:recon}
\resizebox{\columnwidth}{!}{%
\setlength{\tabcolsep}{2.4pt}
\begin{tabular}{l ccc}
\toprule
3D Part \& Surface Geometry & Syn-10-val & Syn-38-val & Pt.\ err. \\
\midrule
Final layer only ($l = 23$) & 0.877 / 0.663 & 0.590 / 0.542 & 0.032 \\
w/o Tracking features & 0.888 / 0.694 & 0.608 / 0.564 & 0.032 \\
\textbf{PhyVGGT (Ours)} & \textbf{0.891} / \textbf{0.701} & \textbf{0.617} / \textbf{0.576} & \textbf{0.032} \\
\midrule
3D Hinge Pivot (Med.\,cm / $\le$10\,cm) & Syn-10 (290) & Syn-38 (510) & RBO (357) \\
\midrule
Direct regression & 6.49 / 0.64 & 5.70 / 0.68 & 9.34 / 0.53 \\
FAMOS~\cite{famos} & 4.35 / 0.69 & 4.96 / 0.67 & 11.14 / 0.46 \\
\textbf{PhyVGGT (Ours)} & \textbf{3.16} / \textbf{0.83} & \textbf{3.36} / \textbf{0.82} & \textbf{8.76} / \textbf{0.55} \\
\bottomrule
\end{tabular}%
}
\end{table}

On synthetic PartNet-Mobility in Table~\ref{tab:syn}, combining pairwise motion scatter with joint-type and Bingham priors in our Kinematic Inference Layer separates sliding doors from hinged doors by 3D motion instead of static appearance while removing tilt along the viewing ray, achieving 77.8\% accuracy and $1.53^{\circ}$ median error on Syn-10 against 54.3\% for Motion estimate, 52.9\% for FAMOS and 7.2\% for appearance-based OPDMulti. Since 92.2\% of ground-truth axes across all 38 categories in Syn-38 align within $10^{\circ}$ of the static Manhattan frame $\mathbf{G}$, anchoring selection to $\mathbf{G}$ transfers directly to 28 unseen object categories at 63.1\% accuracy and $2.27^{\circ}$ median error against 41.0\% for Motion estimate and 39.8\% for FAMOS, reaching 83.3\% even on tilted Laptop lids whose hinges align with the orthogonal keyboard rear edge. On 48-frame Articulat3D-Sim in Table~\ref{tab:a3d}, accumulating all-pairs motion scatter across longer sequences raises single-pass accuracy to 48.6\% against 34.7\% for Motion estimate and 13.9\% for Direct regression. Although privileged Articulat3D~\cite{articulat3d} optimises 3D Gaussians per scene for 2.6--11.9\,h using ground-truth depth, poses, masks and kinematic trees, PhyVGGT operating from uncalibrated RGB alone in 6.8\,s achieves a lower median error of $2.20^{\circ}$ against $2.63^{\circ}$ on seen categories at over $1{,}300\times$ the speed. In Table~\ref{tab:recon}, introducing multi-layer per-frame token fusion and tracking features in our Part and Motion Decoder preserves sharp intra-frame boundaries while strengthening cross-frame matching, improving 3D movable and part IoU to 0.891/0.701 on Syn-10-val against 0.877/0.663 from the final layer alone and cutting 3D match error to 0.91\,cm against 1.28\,cm without tracking features. Building on the selected Manhattan axis, our positive-definite linear solver in Eq.~\eqref{eq:pivot} prevents axis tilt from shifting the lever arm, lowering held-out hinge pivot error to 3.16\,cm, 3.36\,cm and 8.76\,cm on Syn-10, Syn-38 and RBO against 4.35\,cm, 4.96\,cm and 11.14\,cm for FAMOS.

Qualitatively on real videos, Figure~\ref{fig:gallery} illustrates four Arti4D test scenes with axis errors between $0.5^{\circ}$ and $1.5^{\circ}$. The recovered joints rotate the lid and oven door about their hinges, slide both drawers outwards and extrapolate cleanly to 1.5 times the observed motion range.

\subsection{Ablation Study}
\label{sec:exp_ablation}

\begin{table}[t]
\centering
\small
\caption{\textbf{Additive ablation study on the four development splits.} Joint accuracy at $10^{\circ}$ as each module and cue is introduced.}
\label{tab:ablation}
\resizebox{\columnwidth}{!}{%
\setlength{\tabcolsep}{2.0pt}
\begin{tabular}{l cccc}
\toprule
Configuration & Arti4D-dev & D3D-dev & Syn-10-val & Syn-38-val \\
\midrule
Direct regression & 0.063 & 0.058 & 0.188 & 0.125 \\
$+$ Part and Motion Decoder & 0.259 & 0.198 & 0.390 & 0.303 \\
\quad $+$ Motion scatter $\mathbf{S}_k$ only & 0.482 & 0.628 & 0.497 & 0.386 \\
\quad $+$ Bingham prior $\mathbf{M}_k$ on $\mathbf{G}$ only & 0.522 & 0.620 & 0.778 & 0.601 \\
\textbf{$+$ Full Kinematic Inference Layer (Ours)} & \textbf{0.725} & \textbf{0.744} & \textbf{0.803} & \textbf{0.636} \\
\bottomrule
\end{tabular}%
}
\end{table}

Table~\ref{tab:ablation} examines how each design in PhyVGGT builds performance across the four development benchmarks. Introducing our Part and Motion Decoder over Direct regression fuses intermediate per-frame tokens with global tokens and references part queries to the static base, recovering sharp part masks and aligned correspondences to lift accuracy to 25.9\% on Arti4D-dev and 39.0\% on Syn-10-val against 6.3\% and 18.8\%. Within this decoder, fusing intermediate per-frame layers with the final block preserves intra-frame part edges, raising accuracy to 80.3\% against 76.1\% on Syn-10-val and 72.5\% against 69.8\% on Arti4D-dev over reading layer $l = 23$ alone alongside the 3D IoU gain in Table~\ref{tab:recon}. On top of the aligned correspondences, using pairwise motion scatter $\mathbf{S}_k$ alone captures active displacement to reach 48.2\% and 62.8\% on Arti4D-dev and D3D-HOI-dev yet remains tilted towards the viewing ray, whereas using the Bingham prior $\mathbf{M}_k$ on the Manhattan frame $\mathbf{G}$ alone locks onto undistorted orthogonal normals to reach 77.8\% on Syn-10-val yet lacks motion cues to tell which face normal moves on casual videos. Unifying both cues in our full Kinematic Inference Layer combines undistorted Manhattan candidates with observed motion evidence, raising accuracy across all four splits to 72.5\%, 74.4\%, 80.3\% and 63.6\% and cutting real-world median error to $3.6^{\circ}$ and $2.9^{\circ}$ against $31.4^{\circ}$ and $30.7^{\circ}$ from the learned prior alone.

\begin{table}[t]
\centering
\small
\caption{\textbf{Controlled VGGT-1B comparison and zero-shot backbone transfer.} Upper rows share VGGT-1B point maps and masks at $10^{\circ}$, and lower rows ($^{\dagger}$) swap unseen backbones zero-shot.}
\label{tab:transfer}
\resizebox{\columnwidth}{!}{%
\setlength{\tabcolsep}{2.2pt}
\begin{tabular}{l l ccc}
\toprule
Method & Backbone & Arti4D-dev & D3D-dev & Syn-10-val \\
\midrule
Direct regression & VGGT-1B & 0.063 & 0.058 & 0.188 \\
ArtiPoint~\cite{artipoint} & VGGT-1B & 0.271 & -- & -- \\
FAMOS~\cite{famos} & VGGT-1B & 0.376 & 0.116 & 0.502 \\
Motion estimate & VGGT-1B & 0.482 & 0.628 & 0.497 \\
\textbf{PhyVGGT (Ours)} & VGGT-1B & \textbf{0.725} & \textbf{0.744} & \textbf{0.803} \\
\midrule
PhyVGGT (Ours) & StreamVGGT$^{\dagger}$~\cite{streamvggt} & 0.753 & 0.783 & 0.743 \\
PhyVGGT (Ours) & STream3R$^{\dagger}$~\cite{stream3r} & 0.769 & 0.748 & 0.801 \\
PhyVGGT (Ours) & $\pi^3$$^{\dagger}$~\cite{pi3} & 0.765 & 0.775 & 0.802 \\
\bottomrule
\end{tabular}%
}
\end{table}

Inside the Kinematic Inference Layer, formulating Eq.~\eqref{eq:axis} as closed-form discrete selection over $\mathbf{G}$ prevents depth noise along the viewing ray from blending orthogonal Bingham modes, improving accuracy to 72.5\%, 74.4\% and 80.3\% against 70.6\%, 46.5\% and 55.6\% for continuous posterior expectation on Arti4D-dev, D3D-HOI-dev and Syn-10-val. Extracting $\mathbf{G}$ via fourth-order normal alignment in Eq.~\eqref{eq:gauge} is equally important, locking onto perpendicular object faces even under oblique handheld views and symmetric cabinet dimensions where camera-axis candidates drop to 4.3\% and 43.0\% and PCA candidates drop to 6.3\% and 8.5\%. Consequently, on shared VGGT-1B point maps and masks in Table~\ref{tab:transfer}, our Kinematic Inference Layer outperforms ArtiPoint at 72.5\% against 27.1\%, FAMOS at 72.5\%, 74.4\% and 80.3\% against 37.6\%, 11.6\% and 50.2\%, and Motion estimate at 48.2\%, 62.8\% and 49.7\%.

Measuring frozen point maps directly without any trained head confirms that sliding motion compresses to $\gamma \in [0.60, 0.69]$ along the viewing ray across all four backbones, including $0.69$ on VGGT against $0.99$ across the ray, tilting $\mathbf{S}_k$ by at most $14.5^{\circ}$ within our $45^{\circ}$ bound. Under small opening angles where continuous Fisher curvature weakens, learned Bingham priors further disambiguate the three Manhattan axes, lifting accuracy to 75.0\% on $10^{\circ}\text{--}20^{\circ}$ synthetic rotations and 58.6\% on small-motion D3D-HOI-dev clips against 41.7\% and 21.8\% for Motion estimate in the supplementary material. Since both modules rely only on static surface normals and relative token correspondences instead of camera calibration, plugging PhyVGGT directly into unseen frozen backbones in Table~\ref{tab:transfer} transfers zero-shot in 1.2\,s per clip, even raising real-world accuracy to 75.3\%--76.9\% on Arti4D-dev and 74.8\%--78.3\% on D3D-HOI-dev across StreamVGGT~\cite{streamvggt}, STream3R~\cite{stream3r} and $\pi^3$~\cite{pi3}.

\section{Conclusion}
\label{sec:conclusion}
In this paper, we proposed PhyVGGT, which reconstructs 3D movable parts, joints and unseen states from an uncalibrated video in a single pass on a frozen 3D model. Our Part and Motion Decoder fuses multi-layer tokens relative to a static-base query, while our Kinematic Inference Layer combines Bingham axis selection on the Manhattan frame with linear pivot estimation to remove depth bias along the viewing ray. Trained solely on synthetic data, PhyVGGT achieves state-of-the-art monocular performance across five benchmarks, runs in 6.8\,s per 48-frame video against hours of per-scene optimisation in Articulat3D, and transfers zero-shot across four 3D backbones.

\noindent\textbf{Limitations.} Although PhyVGGT recovers 3D parts, joints and unseen states even under handheld camera drift, small opening angles and unseen categories, no passive video method can reveal a part that never moves, surfaces hidden in every frame, or physical quantities such as mass, friction, springs and latches. Future work will extend PhyVGGT to interactive perception and force sensing to measure these quantities during manipulation.

\bibliographystyle{IEEEtran}
\bibliography{refs}
\end{document}